\documentclass[runningheads]{llncs}
\usepackage[T1]{fontenc}
\usepackage{graphicx}
\usepackage{amsmath,amssymb}
\usepackage{varwidth}
\usepackage{hyperref}
\usepackage{float}
\usepackage{caption}
\usepackage{lipsum}
\usepackage{xspace}
\usepackage{bm}
\newbox{\bigpicturebox}
\usepackage{cite}
\usepackage{xcolor}

\def\threeD{3D\xspace}
\def\twoD{2D\xspace}

\def\REALS{\mathbb{R}}
\newcommand{\norm}[1]{\left\lVert{#1}\right\rVert}

\newcommand\x{\mathbf{x}}
\newcommand\h{\mathbf{h}}
\newcommand\dpoint{\mathbf{d}}
\newcommand\omegavec{\bm{\omega}}
\renewcommand\v{\mathbf{v}}
\renewcommand\u{\mathbf{u}}
\renewcommand\c{\mathbf{c}}
\newcommand\p{\mathbf{p}}
\newcommand\hdir{\hat{\h}}
\newcommand\vdir{\hat{\v}}

\newcommand\radius{r}
\newcommand\R{R}
\newcommand\ang{\theta}
\def\equationname{Eq.}
\def\sectionname{Sec.}

\begin{document}
\def\basetitle{Reparametrization of \threeD CSC Dubins Paths Enabling \twoD Search}
%
%
\author{Ling Xu\inst{1} \and
Yuliy Baryshnikov\inst{2} \and
Cynthia Sung\inst{1}}
\authorrunning{L. Xu et al.}
%
\institute{General Robotics, Automation, Sensing \& Perception Lab,\\
University of Pennsylvania, Philadelphia, PA, USA\\
\email{\{xu5,crsung\}@seas.upenn.edu}
\and
Department of Mathematics\\
University of Illinois Urbana-Champaign, Urbana, IL, USA\\
\email{ymb@illinois.edu}
}
\title{\basetitle}
\titlerunning{\basetitle}
\maketitle

\begin{abstract}


This paper addresses the Dubins path planning problem for vehicles in 3D space. In particular, we consider the problem of computing CSC paths --  paths that consist of a circular arc (C) followed by a straight  segment (S) followed by a circular arc (C). These paths are useful for vehicles such as fixed-wing aircraft and underwater submersibles that are subject to lower bounds on turn radius. 
We present a new parameterization that reduces  the 3D CSC planning problem to a  search over 2 variables, thus lowering search complexity, while also providing gradients that assist that search. We use these equations with a numerical solver to explore  numbers and types of solutions  computed for a variety of planar and 3D scenarios. Our method successfully computes CSC paths for the large majority of test cases, indicating that it could be useful for future generation of robust, efficient curvature-constrained trajectories.

\end{abstract}
\section{Introduction} \label{sec:intro}

Aircraft and underwater vehicles are often subject to control constraints, limiting their ability to make sharp turns. These vehicles require robust  planning algorithms to successfully navigate their environments, avoid obstacles, and reach their goal destinations.
The Dubins vehicle model (where vehicles have a lower bound on turn radius) is a well-studied model 
for car-like vehicles in 2D.
A natural extension is to apply these models to \threeD vehicles.
Dubins paths enhance the efficiency and safety of robotic systems in areas including logistics~\cite{huang2023dubins}, surveillance~\cite{faigl2017unsupervised}, search and rescue operations~\cite{ismail2018generating}, and even in robot design~\cite{chen2023kinegami,feshbach2024algorithmic}. Thus, understanding and optimizing Dubins paths hold immense practical importance in modern robotics research and applications.
However, it is challenging to accurately classify the solution space for the Dubins problem in \threeD.




\subsubsection{Background on the \threeD Dubins Path Planning Problem}\label{sec:problemdef}

In this paper, we investigate a solution method for computing \threeD Dubins paths of the CSC type.
Consider a vehicle moving in \threeD space.
Let $\x\in\REALS^3$ be the position of the vehicle and $\R\in SO(3)$ be its orientation. Let $\v=\dot{\x}\in\REALS^3$ and $\omegavec\in\REALS^3$ denote the vehicle's linear velocity and angular velocity, respectively.
%
%
%
For a fixed-wing aircraft~\cite{ambrosino2009path,hota2010,wang2021towards,indig2016near,cui2017reactiveDubins3D,marino2012motion} or underwater drone or submersible~\cite{{9791415,cui2017reactiveDubins3D}} performing low-dynamic maneuvers and simple navigation and waypoint route-planning, a simple model of the vehicle's motion is the \threeD Dubins model: 
\begin{equation}
\left\{\begin{aligned}
    \dot{\v}&=\omegavec\times \v\\
    \omegavec&=\u, \norm{\u}\leq u_{max}.
    \end{aligned}
\right.
\label{eq:dubins}
\end{equation}
That is to say, the vehicle moves at constant forward speed, and the control input $\u$ to the system is the angular velocity only, which is subject to a  upper bound in magnitude $u_{max}$. Both the system's forward speed $\norm{\v}$ and $u_{max}$ are dictated by the particular system, and, without loss of generality, it is often assumed that $\norm{\v}=1$. This is a full \threeD path-planning problem, where other works~\cite{9197084,9568787,4434966,6825891,6842268,patsko2023three,wu2023adaptive,blevins2023real} address a different mathematical problem by modeling fixed-wing unmanned aerial systems through decoupling  \threeD path-planning into a \twoD problem and elevation with a climb rate constraint.
Note that for our simplified \threeD model, the full orientation $\R$ of the vehicle is not important since the angular velocity $\omegavec$ can be specified arbitrarily as long as it satisfies the upper bound constraint.\footnote{In a real vehicle, the elements of $\omegavec$ would be subject to constraints from the vehicle design (e.g, the propeller configuration of an underwater vehicle) and the full orientation $\R$ would be required to relate  each control input to $\omegavec$. We make the assumption that vehicle is equally able to turn in any direction for simplicity. Other works~\cite{9197084,9568787,4434966,6825891,6842268,wu2023adaptive,9791415} have addressed models where directionality matters.}
Thus, the system's current state can be fully characterized by $\x$ and $\vdir$, where  $\hat{\cdot}$ indicates the normalized vector.

Let $\x_i,\vdir_i$ be the starting position and orientation of a \threeD vehicle, and $\x_f,\vdir_f$  its goal position and orientation. The \emph{Dubins path planning problem} is:

\begin{problem}[\threeD Dubins Planning]
Given an initial configuration $(\x_i,\vdir_i)$, a final configuration $(\x_f,\vdir_f)$, and a control input upper bound $u_{max}$, find the shortest-length path between points $\x_i$ and $\x_f$ such that the initial and final directions are given by $\vdir_i$ and $\vdir_f$, respectively, subject to \equationname~\eqref{eq:dubins}.
    \label{pr:dubins}
\end{problem}
%
%
%
%
%
%
%
Since the vehicle velocity is constant, this problem can be equivalently written as a geometric problem of finding a shortest length path with a curvature constraint:

\begin{problem}
Given an initial configuration $(\x_i,\vdir_i)$, a final configuration $(\x_f,\vdir_f)$, and a control input upper bound $u_{max}$, find the shortest-length path between points $\x_i$ and $\x_f$ such that the initial and final directions are given by $\vdir_i$ and $\vdir_f$ and the path curvature does not exceed $1/u_{max}$.
    \label{pr:dubinsgeo}
\end{problem}

This model is an extension to the \twoD classical Dubins  model first introduced in 1957 in~\cite{dubins1957curves} for a car-like vehicle that 
can only move forward in the plane at constant speed with bounded  minimum turn radius,
and various approaches exist to find optimal paths. One common method involves the application of the Pontryagin's Maximum Principle, which provides necessary conditions for an optimal control problem \cite{sachkov2022left}. 
This principle implies that the optimal trajectories solving this problem are of bang-bang type (arcs of the maximal possible curvature) connected, possibly, by sliding trajectories. A careful analysis reveals that the sliding trajectories need to be straight lines \cite{shkel2001classification, 220117}. Further, the number of fragments is at most three.
Dubins~\cite{dubins1957curves} proved that optimal trajectories must necessarily fall into one of two forms: CSC or CCC, where C denotes a circular arc of radius $1/u_{max}$ and S  a straight line segment. For any initial and final configuration, the Dubins set is defined to be the six admissible paths LSL, RSR, RSL, LSR, RLR, and LRL, with L denoting a left turn and R a right turn.

%
%
When the problem is extended to \threeD, trajectories must still  be of bang-bang type, but the optimal trajectories now consist of helicoidal trajectories (spiraling around a cylinder with constant gain), including a special case when the gain is zero, i.e., the trajectory is planar (an arc). As in the planar case, sliding regimes are possible and, again, can be shown to consist of straight lines.
One can ask whether the helicoidal trajectories are truly necessary. To this end, 
\cite{sussmann1995shortest} showed that there exist pairs of initial and final configurations such that helicoidal trajectories are strictly shorter than any CSC or CCC paths, and \cite{6825891} constructed these \threeD helical paths.
For many practical purposes, the CSC or CCC paths are far more desirable since they are easier to parameterize geometrically and easier to implement in practice, as each of the fragments is planar.

In this paper, we focus on an efficient algorithm for computing CSC trajectories connecting two \threeD configurations and satisfying
\equationname~\ref{eq:dubins}:
\begin{problem}[\threeD CSC Paths]
    Given an initial configuration $(\x_i,\vdir_i)$, a final configuration $(\x_f,\vdir_f)$, and a minimum radius of curvature $\radius$, find valid CSC paths between points $\x_i$ and $\x_f$ such that the initial and final directions are given by $\vdir_i$ and $\vdir_f$ and the path curvature does not exceed $1/\radius$.
\end{problem}
Existing approaches often face limitations that restrict their generality.
Hota and Ghose~\cite{hota2010} provide a geometric formulation for finding \threeD CSC Dubins paths that requires numerically solving a nonlinear system of equations with five unknown variables, 
but their solution only applies when starting and ending points are sufficiently far apart. Further, they did not provide an analysis of how the number of solutions vary with the \threeD configuration space. 
Some \threeD path planning algorithms rely on optimal control methods~\cite{4434966, ambrosino2009path,wang2021towards}. While they are computationally tractable, they do not admit an analytical solution and are only resolution complete. Other numerical methods involving iterative procedures such as optimization for a non-linear programming formulation \cite{9568787} and RRT  \cite{6842268} can fail to converge to the optimal solution for certain \threeD cases and require exhaustive search. Recent work in \cite{baez2024analyticCSC} computes up to 7 valid \threeD CSC Dubins paths, but the formulation of the Dubins problem as a 6R manipulator and solving for the analytical solutions of the inverse-kinematics does not address applicability to real-time path-planning that gradient-based numerical solvers enable.


\subsubsection{Contributions}

We present a parametrization of the \threeD CSC Dubins problem that simplifies computation of an optimal path, reducing the problem to a numerical search over two variables,  $h_i$ and $h_f$. 
We ignore for now the problem of CCC paths, leaving that for future investigation.
For CSC paths, then, we develop closed-form equations (Section~\ref{sec:Parametrization}) that describe the desired starting and ending configurations as a function of  $h_i$ and $h_f$, thereby allowing one to take gradients of the equations with respect to these parameters. 
This gradient can assist 
a numerical solver when the paths are more geometrically complex. We characterize the solution space of \threeD CSC Dubins paths in both planar and non-planar cases (Section~\ref{sec:results}), and we show how the new formulation improves the detection of \threeD CSC Dubins path solutions compared to an existing work~\cite{hota2010optimal}. 

\section{Parametrization of the \threeD Dubins Planning Problem}\label{sec:Parametrization}

\begin{figure}[tb]
\centering
        \includegraphics[width=\textwidth]{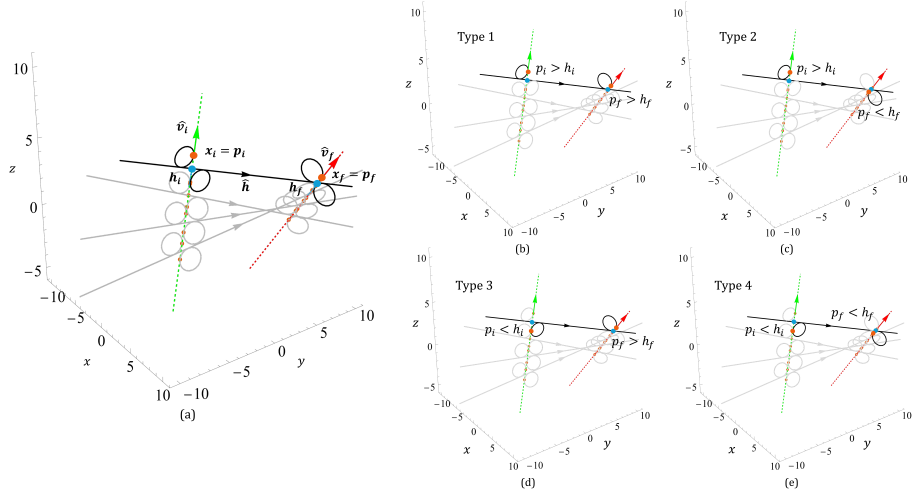}
        \caption{\textbf{Parametrization of \threeD CSC Paths } \textbf{(a)} A straight line in the $\hdir$ direction  intersects with lines through the starting and ending configurations, resulting in two possibilities for the starting point $\p_i$ and two possibilities for ending point $\p_f$. \textbf{(b-e)} Type 1-4 starting and ending circles.}
\label{fig1}\label{fig:parameters}
\end{figure}

To start, we observe that regardless the orientations of the initial and final configurations, the initial CS component of the path is planar, and so is the final SC component of the path.
Further, if the line along which the S component lies is known, then the planes containing the two C components are completely defined.
Let us introduce two new points
\begin{equation}
\h_i=\x_i+h_i\vdir_i, \hspace{3mm} \h_f=\x_f+h_f\vdir_f,
    \label{eq:h}
\end{equation}
which lie on the lines defined by the initial and final configurations, respectively.
Assume that the S component of the CSC path lies on the line from $\h_i$ to $\h_f$.
We define $\hdir$ to be the normalized vector corresponding to $\h_f-\h_i$.
The initial C component of the path must lie in the plane defined by $\x_i$ and the vectors $\vdir_i$ and $\hdir$, and similarly for the final C component.
Figure~\ref{fig:parameters} shows in gray lines how this plane and the resulting C components change with $h_i$ and $h_f$.

Consider the initial C component. There are two circles of radius $\radius$ that lie tangent both to the line defined by $\x_i$ and $\vdir_i$ and to the line defined by $\h_i$ and $\h_f$ and correctly oriented for the vehicle to transition from an initial orientation of $\vdir_i$ to a straight-line orientation of $\hdir$. The centers of these circles are
\begin{equation}
    \c_i=\h_i\pm \frac{\radius}{\norm{\vdir_i\times\hdir}}\left(\vdir_i-\hdir\right)
    \label{eq:c}
\end{equation}
We will use $\c_{i1}$ to indicate the $+$ solution and $\c_{i2}$ to indicate the $-$ solution. 
The point of tangency between the circle and the line defined by $(\x_i,\vdir_i)$ can then be computed as the projection of $\c_i$ onto the line:

\begin{equation}
    \p_i=\x_i+p_i\vdir_i, \hspace{3mm} p_i=h_i \pm \frac{\radius}{\norm{\vdir_i\times\hdir}}\left(1-\hdir\cdot\vdir_i\right)
    \label{eq:px}
\end{equation}
Again, we will use $p_{i1}$ (and equivalently, $\p_{i1}$) to indicate the $+$ solution and $p_{i2}$ ($\p_{i2}$) to indicate the $-$ solution.
The same analysis can be conducted for the final C component, where the center is defined as $\c_f$, resulting in:
\begin{equation}
        \p_f=\x_f+p_f\vdir_f, \hspace{3mm} p_f=h_f\pm \frac{\radius}{\norm{\vdir_f\times\hdir}}\left(1-\hdir\cdot\vdir_f\right)\\
    \label{eq:pf}
\end{equation}
The points $\h_i$ and $\h_f$ correctly identify the desired CSC path if one of $\p_{i1}$ or $\p_{i2}$ is $\x_i$ and one of $\p_{f1}$ or $\p_{f2}$ is $\x_f$ such that the scalar quantities satisfy:
\begin{equation}
\left\{
\begin{array}{ccc}
    p_{i1}=0   &\text{\ \ \ or\ \ \ }    &p_{i2} = 0\\
    p_{f1}=0   &\text{\ \ \ or\ \ \ }    &p_{f2} = 0
\end{array}\right.
\end{equation}
We categorize these solution types in \tablename~\ref{tab:types} with the geometries shown in \figurename~\ref{fig:parameters}.

\begin{table}[b]
    \centering
    \caption{Solution Types}
    \renewcommand{\arraystretch}{1.2}
    \setlength{\tabcolsep}{5pt}
    {\scriptsize
    \begin{tabular}{c|c|c|}
    \multicolumn{3}{c}{\textbf{Regular}}\\
        & $p_{i1}=0$ & $p_{i2}=0$\\\hline
         $p_{f1}=0$& Type 1  & Type 3 \\\hline
         $p_{f2}=0$&  Type 2& Type 4\\\hline
    \end{tabular}}
    \hfil
    {\scriptsize
    \begin{tabular}{c|c|c|}
    \multicolumn{3}{c}{\textbf{Switched}}\\
        & $p^*_{i1}=0$ & $p^*_{i2}=0$\\\hline
         $p^*_{f1}=0$& Type 5  & Type 7 \\\hline
         $p^*_{f2}=0$&  Type 6& Type 8\\\hline
    \end{tabular}
    }
    \label{tab:types}
\end{table}

\subsubsection{Path Directionality} 
\label{sec:invalidsols}
While the given equations ensure that the directions of circular arc components are consistent with the directions of the given $\hdir$ vector, it is possible for  the intersection between the starting circle centered around $\c_i$ and the line from $\h_i$ to $\h_f$ to occur ``before'' the intersection between the ending circle centered around $\c_f$ and the line. For this situation, the vehicle would  have to travel in the reverse direction along the straight line component, from $\h_f$ to $\h_i$. 
We therefore consider four additional solution types, corresponding to ``switched'' situations in which the vehicle must travel along the straight line segment in the reverse direction. 
\begin{figure}[tb]
    \centering
    \includegraphics[width=0.97\textwidth]{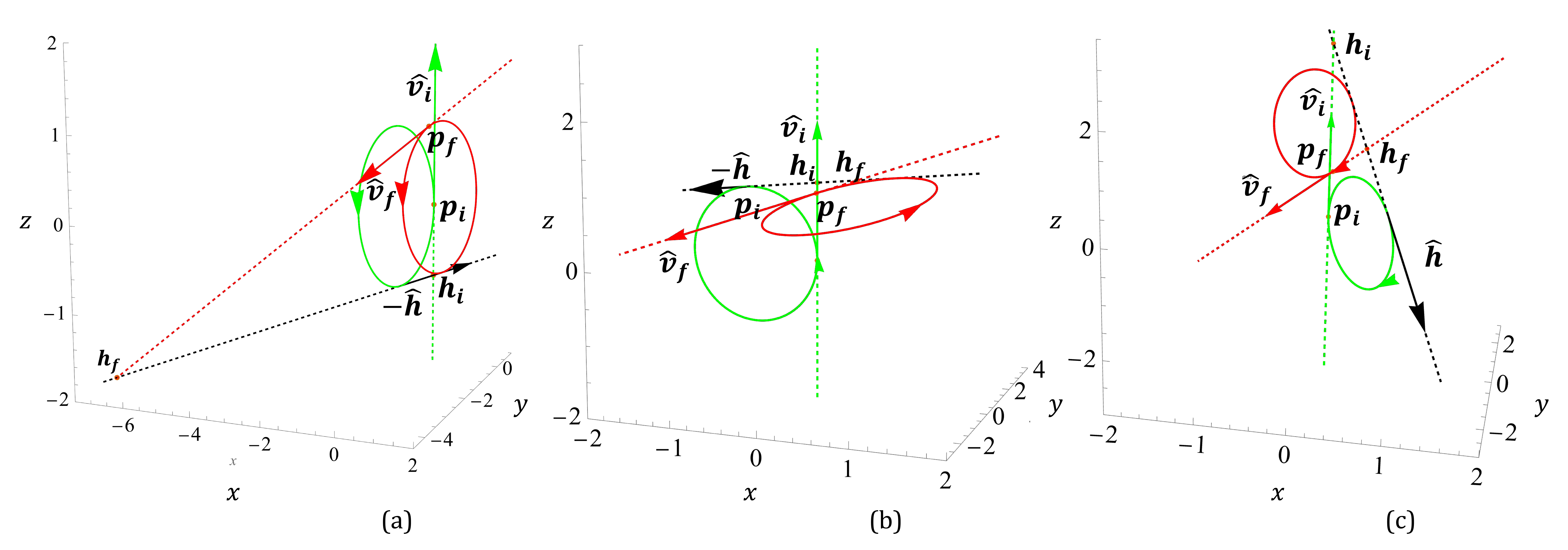}
    \caption{ \textbf{Path Directionality} \textbf{(a)} Example of a ``switched'' solution where the straight line path is traveling along the $-\hdir$ direction. \textbf{(b)} Example of an invalid switched-type computed path  in \threeD where the $-\hdir$ direction travels away from the ending point. \textbf{(c)} Example of a 
    an invalid regular-type computed path in \threeD where  the $\hdir$ direction travels away from the ending point.}
    \label{fig:switched}
\end{figure}
These paths can be computed using the same equations as for the regular case, except replacing $\hdir$ with $-\hdir$. That is, 
\begin{equation}
        p_i^*=h_i \pm \frac{\radius}{\norm{\vdir_i\times-\hdir}}\left(1+\hdir\cdot\vdir_i\right),\hspace{1cm}
        \p_i^*=\x_i+p_i\vdir_i
\label{eq:p}
\end{equation}
\begin{equation}
    p_f^*=h_f\pm \frac{\radius}{\norm{\vdir_f\times-\hdir}}\left(1+\hdir\cdot\vdir_f\right),\hspace{1cm}
    \p_f^*=\x_f+p_f\vdir_f.
    \label{eq:pf_switch}
\end{equation}
We still use $p^*_{i1}$ (and equivalently, $\p^*_{i1}$) to indicate the $+$ solution and $p^*_{i2}$ ($\p^*_{i2}$) to indicate the $-$ solution, and similarly for $p^*_f$ and $\p^*_f$.


Similarly to the regular situation, when the scalar quantities satisfy
\begin{equation}
\left\{
\begin{array}{ccc}
    p^*_{i1}=0   &\text{\ \ \ or\ \ \ }    &p^*_{i2} = 0\\
    p^*_{f1}=0   &\text{\ \ \ or\ \ \ }    &p^*_{f2} = 0
\end{array}\right.
\end{equation} we have the desired CSC path. 
We categorize these solution types in \tablename~\ref{tab:types}.

\subsubsection{Full Solution Set}
Thus, for any given initial and final configuration, there are $8$ possible solution types. 
It is not necessarily the case that every solution type yields a solution, and in some cases, there are multiple unique paths of one solution type.
Additionally,  it is possible for the solution for $h_i$ and $h_f$ to be invalid even if the equations are satisfied. In particular, the direction of the straight line path, either $\hdir$ or $-\hdir$, may be in the opposite direction of the desired ending point. 
%
Invalid paths can be identified as follows:
\begin{enumerate}
    \item {Regular solutions} invalid when  $(\c_f-\c_i) \cdot \hdir < 0$ as shown in \figurename~\ref{fig:switched}(b). 
    \item {Switched solutions} invalid when  $(\c_f-\c_i) \cdot \hdir > 0$ as shown in \figurename~\ref{fig:switched}(c). 
\end{enumerate} 

\subsection{Extracting the CSC Path}
The components of the resulting CSC path can be extracted from $h_i$ and $h_f$. 
For the initial C component, the plane on which the arc lies is defined by the vectors $\vdir_i$ and $\hdir$.
The center of the circular arc can be computed according to \equationname~\ref{eq:c}.
The arc length $\ang_i$ is given by
\begin{equation}
    \ang_i=\left\{\begin{array}{cc}
    \cos^{-1}{\left(\vdir_i\cdot\hdir\right)}, & p_i<h_i \\
    2\pi-\cos^{-1}{\left(\vdir_i\cdot\hdir\right)},& p_i>h_i
    \end{array}\right.
\end{equation}
These same equations also hold for the final circular arc.

The S component consists of the line between $\h_i$ and $\h_f$ where it intersects with the initial and final arc. 
%
%
%
For the initial arc, the intersection $\dpoint_i \in \mathbb{R}^3$ 
can be found
by projecting the center of the circle  onto the vector $\hdir$:
\begin{equation}
    \dpoint_i = \left[(\c_i - \h_i) \cdot \hdir \right]\hdir + \h_i 
\end{equation}
and similarly for the final arc.
Then the S component is the straight line path from $\dpoint_i$ to $\dpoint_f$. This segment is in the $\hdir$ direction for regular type paths and in the $-\hdir$ direction for switched type paths. The two arcs together with the straight line form the CSC path. 

%


\subsection{Gradients} \label{sec:gradients}

The previous subsections have demonstrated that it is possible to solve for CSC Dubins paths as a system of two equations on two scalar variables $h_i$ an $h_f$. 
These closed-form equations admit gradients, which allow us to solve them more efficiently using numerical solvers.
%
%
These derivatives take the following form:
\begin{align}
    \frac{\partial p_{i}}{\partial h_i}&= \pm \frac{-r(1 - \hdir \cdot \vdir_i)(\vdir_i \times -\hdir) \cdot (\vdir_i \times - a_i))}{\lVert \vdir_i \times -\hdir \rVert^3} -\frac{r(a_i \cdot \vdir_i)}{\lVert \vdir_i \times  -\hdir \rVert} + 1\\
    \frac{\partial p_{i}}{\partial h_f}&= \pm \frac{-r(1 - \hdir \cdot \vdir_i)(\vdir_i \times -\hdir) \cdot (\vdir_i \times  -a_f)}{\lVert \vdir_i \times -\hdir \rVert^3} \--\frac{r(a_f \cdot \vdir_i)}{\lVert \vdir_i \times -\hdir\rVert}\\
    \frac{\partial p_{f}}{\partial h_i}&= \pm \frac{-r(1 - \hdir \cdot \vdir_f)(\vdir_f \times -\hdir) \cdot (\vdir_f \times -a_i)}{\lVert \vdir_f \times -\hdir \rVert^3} -\frac{r(a_i \cdot \vdir_f)}{\lVert \vdir_f \times -\hdir\rVert} + 1\\
    \frac{\partial p_{f}}{\partial h_f}&= \pm \frac{-r(1 - \hdir \cdot \vdir_f)(\vdir_f \times -\hdir) \cdot (\vdir_f \times -a_f))}{\lVert \vdir_f \times -\hdir \rVert^3} -\frac{r(a_f \cdot \vdir_f)}{\lVert \vdir_f \times -\hdir \rVert}
\end{align}
where
\begin{equation}
    a_i = \frac{d\hdir}{dh_i} = \frac{(\hdir \cdot \vdir_i)\hdir - \vdir_i}{\lVert \h_f - \h_i \rVert}, \hspace{3mm}
    a_f = \frac{d\hdir}{dh_f} = \frac{-(\hdir \cdot \vdir_f)\hdir + \vdir_f}{\lVert \h_f - \h_i \rVert}.
\end{equation}
The $+$ in the equations corresponds to derivatives of $p_{i1}$ or $p_{f1}$ with respect to $h_i$ and $h_f$, and the $-$ corresponds to derivatives of $p_{i2}$ or $p_{f2}$ with respect to $h_i$ and $h_f$.
Derivatives for switched equations $p^{*}_i$ and $p^{*}_f$ are identical except for replacing $\hdir$ with $-\hdir$.

\section{Experimental Results}\label{sec:results}

Given initial configuration $\x_i, \vdir_i$ and final configuration $\x_f, \vdir_f$, we set $\x_i = \p_i$ and $\x_f = \p_f$ and solve for $h_i$ and $h_f$ for each solution type. We implement a CSC path solver in Python using \texttt{scipy.optimize.fsolve} to solve the equations in \sectionname~\ref{sec:Parametrization}. Tests were conducted on an 8-core AMD Ryzen 7 5825U processor with 16 GB RAM. 
Solutions for each pair of equations (Types 1-8) are computed 
over the variables $h_i$ and $h_f$ and using the gradients given in \sectionname~\ref{sec:gradients}.
We filter out invalid solutions that may stem from either failure of the numerical solver to converge (which may be detected by checking the values of $p_i$ and $p_f$) or from directional inconsistencies
as mentioned in \sectionname~\ref{sec:invalidsols}. We test the effectiveness and versatility of these 8 equations in detecting possible CSC paths for both non-planar and planar cases. All of the tests are conducted with $r=1$. 


\subsection{Example Solutions}\label{sec:results_examples}

We begin by testing specific planar and non-planar cases to explore the solver's ability to compute different solution types. For both planar and non-planar cases, the difficulty of the problem seems to depend on the proximity of the starting and ending positions. Additional examples can be found in Appendix~\ref{app:examples} 

\begin{figure}[tb]
\hspace*{-1mm}\includegraphics[width=0.95\textwidth]{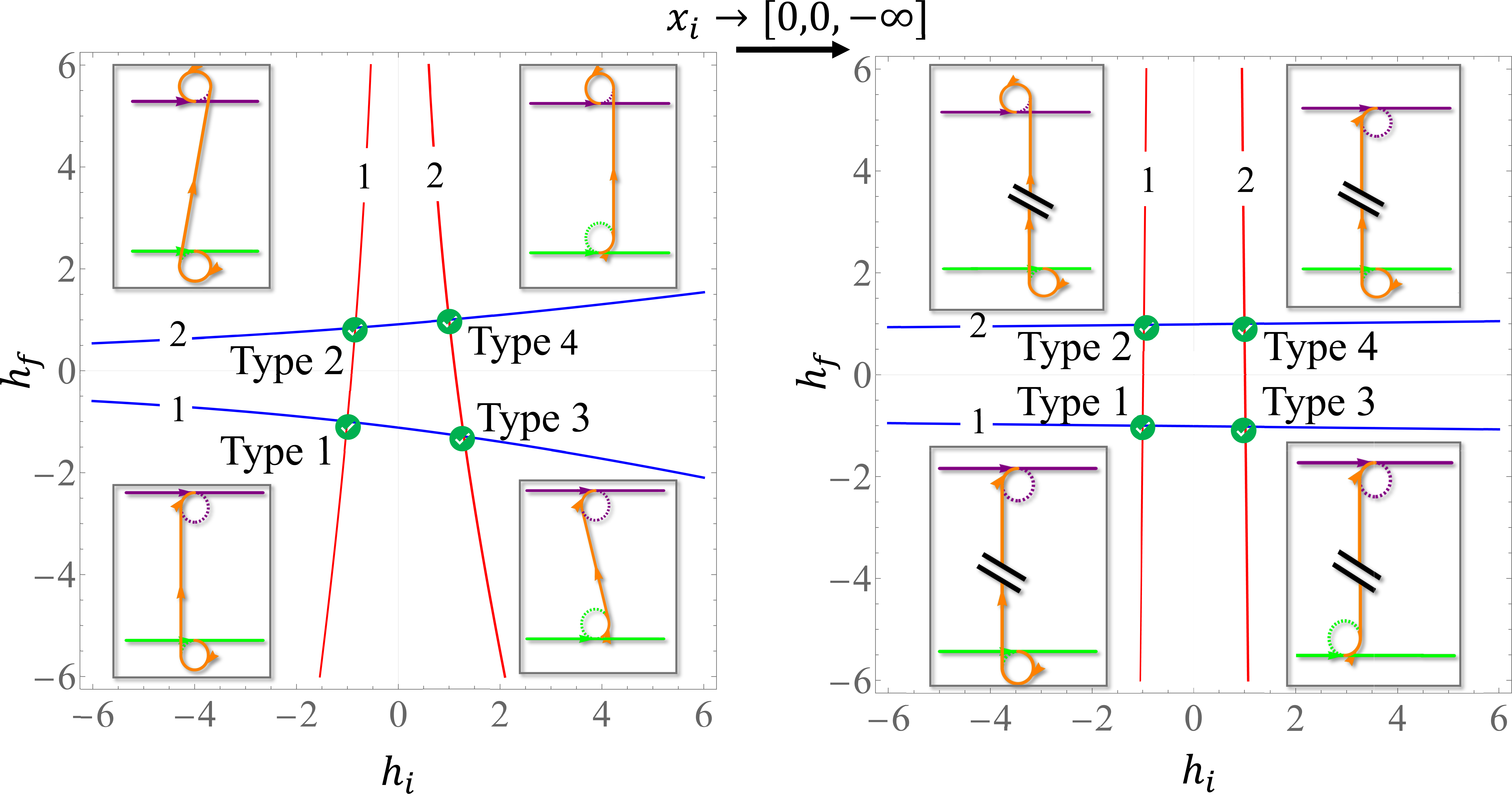}%
    \caption{\textbf{Solution plots}:  $\x_f = [0,0,0], \vdir_i = [1,0,0] = \vdir_f$.  $\x_i = [0,0,-10]$ (left) vs $\x_i = [0,0 -100]$ (right). Values for $h_i$ and $h_f$ correspond to $p_i=0$ (red) and $p_f=0$ (blue). The (1) annotation indicates $p_{i1}$ and $p_{f1}$ curves and (2) marks $p_{i2}$ and $p_{f2}$ curves. Intersections marked by green checks  are valid solutions. Insets in 4 corners show the solution types, with  hashes indicating cutting out a portion of the path due to points being far apart. Green and purple lines indicate the starting and ending directions, respectively. The final CSC path is in orange.} 
    \label{fig:faraway_contour}
\end{figure}

\subsubsection{Planar Cases}

We begin by examining the planar case for which methods for computing solutions are well-established.
When the starting and ending points are placed sufficiently far apart, then all four types of regular solutions exist and the equations for $p_i$ and $p_f$ are fairly well-behaved.
%
Figure~\ref{fig:faraway_contour} shows the solutions for $\x_i \rightarrow [0,0,-\infty]$ and $\x_f = [0,0,0]$. Values for $h_i$ and $h_f$ corresponding to $p_i=0$ are shown in red, and values corresponding to $p_f=0$ are shown in blue. 

\begin{figure}[tb]
\centering
    \includegraphics[width=0.84\textwidth]{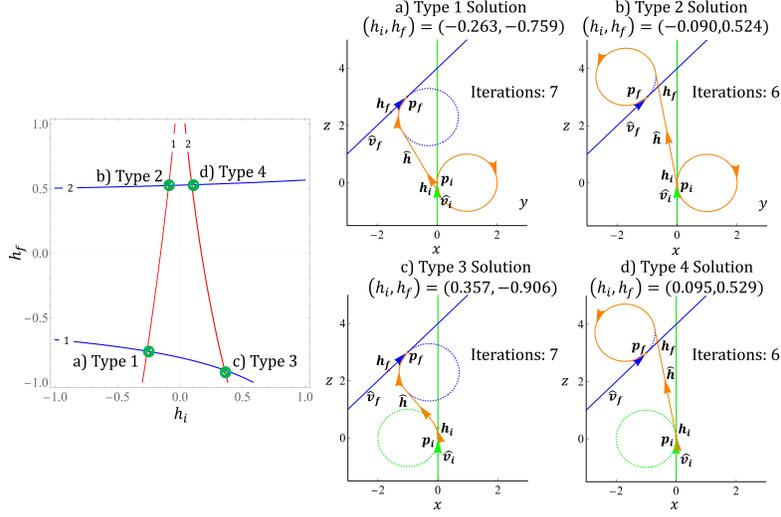}   
\caption{\textbf{\twoD  Problem Case 1 (Far)}: $\x_i = [0,0,0], \x_f = [-1, 0, 3],  \vdir_i = [0,0,1], \vdir_f = \frac{1}{\sqrt{2}}{[1,0,1]}$. (Left) Solution plot for regular solutions of $p_i$ (red) and $p_f$ (blue). Intersections marked by green checks show the $(h_i, h_f)$ solution pairs. (Right) CSC paths for each solution type 1-4 marked on the left. 
}
 \label{fig:planarfar1}
\end{figure}

All of the curves are smooth and approach horizontal ($p_f = 0$) and vertical lines ($p_i$ = 0) with straightforward intersections representing candidate solutions as $\x_i \rightarrow [0,0,-\infty]$.
This makes sense since as the starting and ending positions move further apart, changes in $h_f$ will have  less of an effect on the position of $p_i$, and similarly for the effect of $h_i$ on $p_f$. We thus expect that solving for CSC paths when starting and ending positions are far to be straightforward for a standard numerical solver.
%
For example,
Figure~\ref{fig:planarfar1} 
shows results for 
one test condition where the starting and ending points are "sufficiently far":
 $\x_i = [0,0,0], \x_f = [-1, 0, 3], \vdir_i = [0,0,1], \vdir_f = \frac{1}{\sqrt{2}}[1,0,1]$. 
Similarly to the previous example, this case produces 4 regular-type solutions, one of each type.
The solutions were all found in 0.038~s 
with fewer than 10 solver iterations for each solution type. 

\begin{figure}[tb]
\centering
    \includegraphics[width=0.84\textwidth]{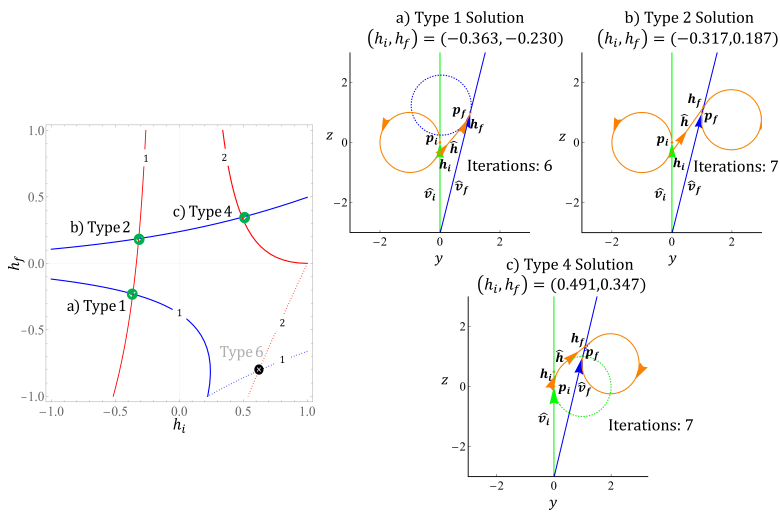}   
    \caption{\textbf{\twoD Problem Case 1 (Close)}: $\x_i = [0,0,0], \x_f = [0, 1.01, 1], \vdir_i = [0,0,1], \vdir_f = \frac{1}{\sqrt{17}}{[0,1,4]}$. (Left) Solution plot for regular solutions of $p_i$ (solid red), regular solutions of $p_f$ (solid blue), switched solutions of $p^*_i$ (dotted red), and switched solutions of $p^*_f$ (dotted blue). 
    Intersections marked by green checks show the $(h_i, h_f)$ valid solution pairs, 
    and
    black xs show invalid solutions.
    (Right) CSC paths for each solution type. A Type 3 solution does not exist in this case.}
   \label{fig:planarclose1}
\end{figure}

More challenging cases occur when the starting and ending positions are close together. In these cases, not all regular-type solutions are valid, or even exist.
Figure~\ref{fig:planarclose1} 
shows the results for one such test condition:
 $\x_i = [0,0,0], \x_f = [0, 1.01, 1], \vdir_i = [0,0,1], \vdir_f = \frac{1}{\sqrt{17}}[0,1,4]$. 
The solver is able to successfully find all valid solutions for this case. We begin to see non-regular type candidate solutions represented by intersections of dotted lines on the plot, and there are only 3 valid solutions since there is no straight line path between an initial right-turning arc and a final left-turning arc.
In addition to these regular solutions, one solution of Type~6 was found but was filtered as having an invalid straight segment direction.
The solver took 0.146~s to find all of the solutions 
due to the need to filter out more solutions that either do not exist or are invalid. 

\begin{figure}[tb]
\centering
      \includegraphics[width=0.84\textwidth]{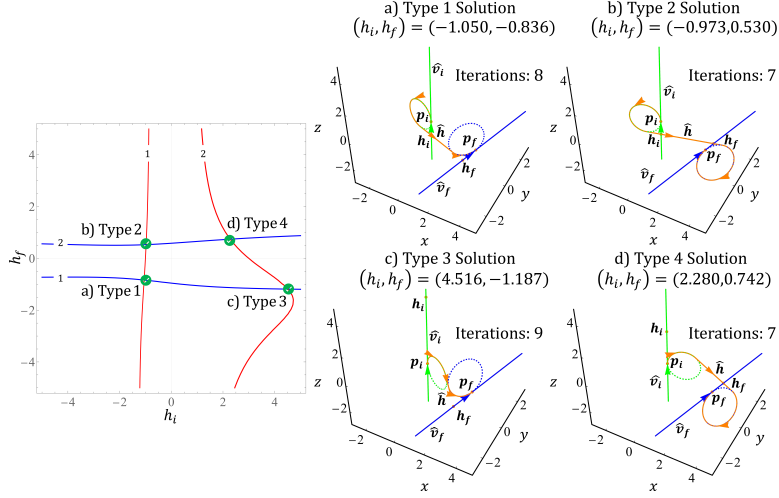}
    \caption{\textbf{\threeD Problem Case 1 (Far)}: $\x_i = [0,0,0], \x_f = [3, 0, -1], \vdir_i = [0,0,1], \vdir_2 = \frac{1}{\sqrt{21}}[2,4,1]$.
     (Left) Solution plot for regular solutions of $p_i$ (red) and $p_f$ (blue). Intersections marked by green checks show the $(h_i, h_f)$ solution pairs. (Right) CSC paths for each solution type 1-4 marked on the left.
    }
\label{fig:3dfar1}
\end{figure}

\subsubsection{Non-Planar Cases} 

The same equations extend to \threeD scenarios, showing similar trends for starting and ending positions that are close or far apart.

Figure~\ref{fig:3dfar1} shows the results for one  \threeD case where the starting and ending positions are sufficiently far apart in distance: 
$\x_i = [0,0,0], \x_f = [3, 0, -1], \vdir_i = [0,0,1], \vdir_2 = \frac{1}{\sqrt{21}}[2,4,1]$.
Similarly to the planar tests, this is fairly easy for the solver, producing all four solutions (one of each regular type) in 0.031~s and requiring fewer than 10 solver iterations for each solution pair.




\begin{figure}[tb]
\centering
    \includegraphics[width=0.85\textwidth]{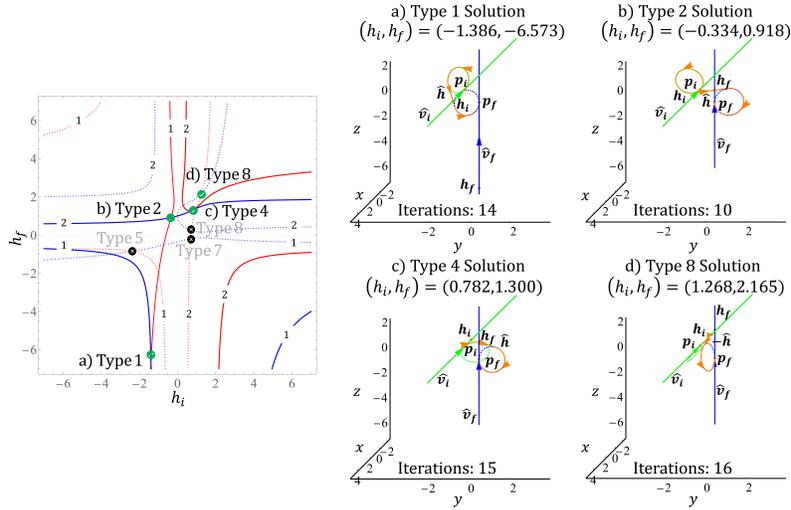}
    \caption{\textbf{\threeD  Problem Case 1 (Close)}: $\x_i = [0,0,0], \x_f = [-1, 0, 3], \vdir_i = \frac{1}{3}[1,1,1], \vdir_f = [0,0,1]$.
     (Left) 
    Solution plot for regular solutions of $p_i$ (solid red), regular solutions of $p_f$ (solid blue), switched solutions of $p^*_i$ (dotted red), and switched solutions of $p^*_f$ (dotted blue). 
    Intersections marked by green checks show the $(h_i, h_f)$ valid solution pairs,
    and
    black xs show invalid solutions. 
    There are 3 regular, 1 switched, and 3 invalid (Type 5, 7, 8) solutions.
    (Right) 
    CSC paths for each of the solution types marked on the left.
    }
   \label{fig:Ng1} \label{fig:3dclose1}
\end{figure}


When starting and ending positions are close together in \threeD, similarly to the planar case, 
there are not only solutions of the four regular types.
Figure~\ref{fig:3dclose1} shows the results for one such example:
$\x_i = [0,0,0], \x_f = [-1, 0, 3], \vdir_i = \frac{1}{\sqrt{3}}[1,1,1], \vdir_f = [0,0,1]$ 
The solver found all intersection points in 0.061~s.
The number of iterations required for the numerical solver to find solutions increases when the starting and ending positions are close together, and some equation pairs produce invalid $(\h_i,\h_f)$ solutions that need to be filtered out.

\subsection{Slices of  Planar and Non-Planar Solution Spaces}
To explore the solution space and the difficulty of computing solution paths, we sweep a range of starting and ending configurations for both \twoD and \threeD cases.

\subsubsection{Planar Solution Spaces}
\begin{figure}[tb]
    \centering
    \hspace*{-1.64mm}\includegraphics[width=0.88\textwidth]{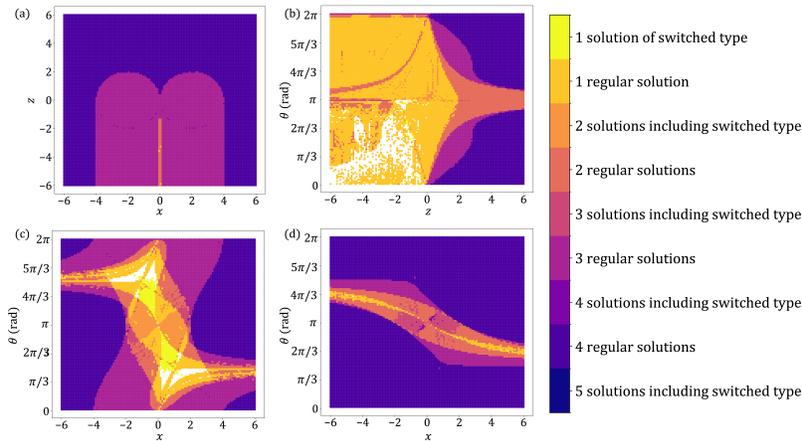}
    \caption{\textbf{Planar Solution Spaces} \textbf{(a)} $x-z$ plane when $\theta = 0$. \textbf{(b)} $z-\theta$ plane when $x = 0$. \textbf{(c-d)} $x-\theta$ plane when $z = -1 $ (left) and $z = 3$ (right).}
    \label{fig:planar_slices}
\end{figure}

Fixing the starting point at $\x_i = [0,0,0]$ and the starting direction at $\vdir_i = [1,0,0]$, we investigate the solution space for planar cases by varying the end position $\x_f = [x,0,z]$ and  direction $\vdir_f = [-\sin(\theta), 0,\cos(\theta)]$ for values of $x, z \in [-6,6]$ and $\theta \in [0,2\pi)$. The initial guess for the solver is set at $h_i = h_f = 0$.
Figure~\ref{fig:planar_slices} shows the results. Slices of the problem space are colored according to the number of solutions (regular and switched) found.
The solver is able to find at least one solution for most inputs. Further, the solver produces expected symmetries.
For example, as shown \ref{fig:planar_slices}(a), when $\theta = 0$, the starting and ending directions are aligned, so we expect that the number of solutions should be symmetric around $x = 0$.
If $x = 0$, the number of solutions should be symmetric about $\theta = \pi$, shown in Figure \ref{fig:planar_slices}(b).
Finally, when fixing the $z$ coordinate in $\x_f = [x,0,z]$, the number of solutions should be symmetric about $x = 0$ and $\theta = \pi$, as shown in the Figure \ref{fig:planar_slices}(c).

Similarly to the qualitative observations on our individual test cases, the plots reveal that when the starting and ending positions are far apart, the solver is able to find 4 regular solutions much of the time. However, as the distance between starting and ending position decreases, the number of regular solutions also decreases, and switched solutions must be found. We did not compute the true number of solutions for each of the tested scenarios. It remains to be determined for some of the cases where fewer solutions were found whether it is because there are fewer solutions or because the solver did not find them all. 

\subsubsection{Non-Planar Solution Spaces}
\begin{figure}[tb]
    \centering
    \includegraphics[width=0.88\textwidth]{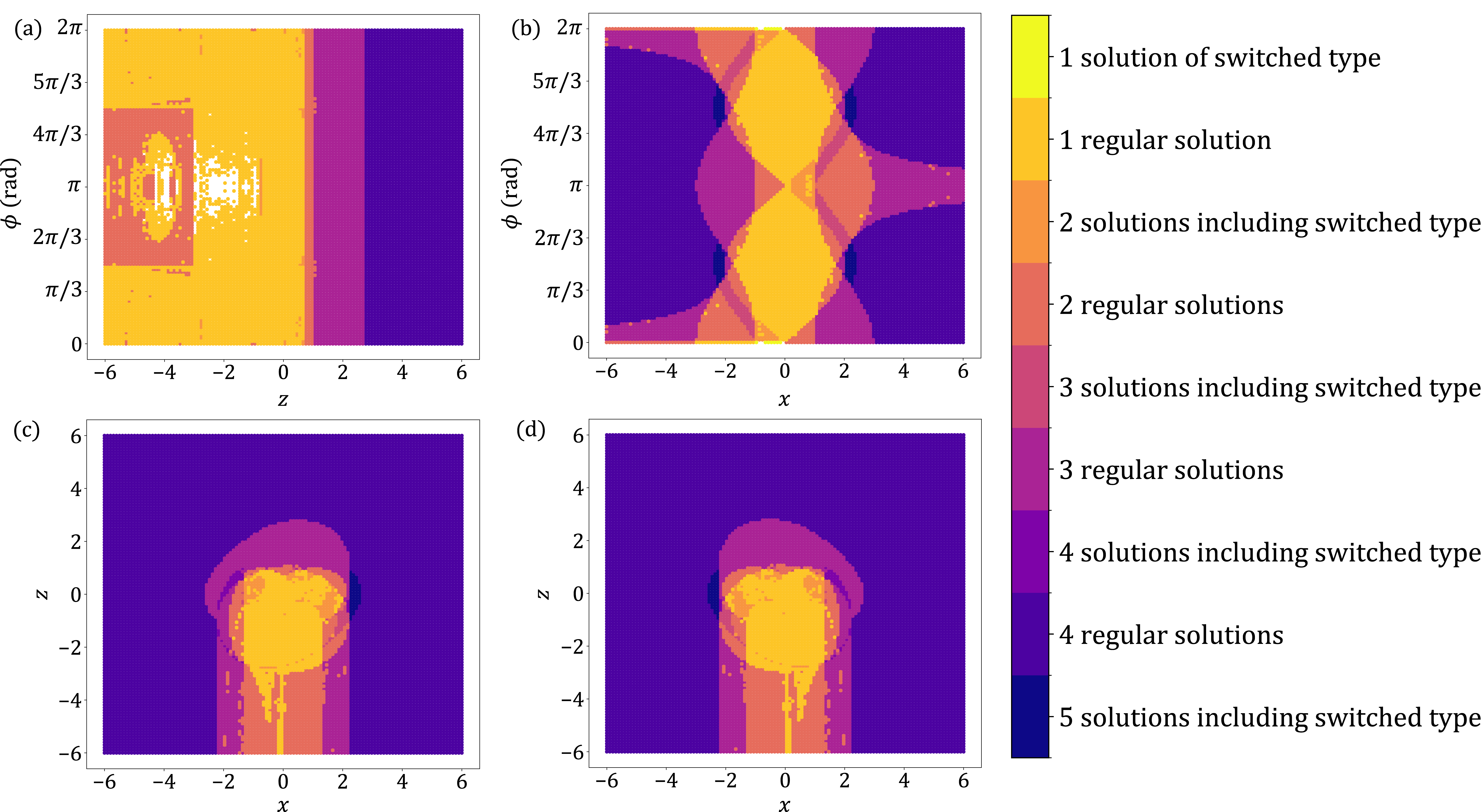}
    \caption{\textbf{Non-planar Solution Spaces} 
 \textbf{(a)} $z-\phi$ plane when $x = 0$. \textbf{(b)} $x-\phi$ plane when $z = -1$. 
    \textbf{(c-d)} $x-z$ plane slice when $\phi = \pi - 2$ rad (left) and and $x-z$ plane slice when $\phi = 2$ rad (right).}
    \label{fig3}\label{fig:3dsweep}
\end{figure}

Extending to non-planar problems, we consider the following situation.
Fixing the starting point at $\x_i = [0,0,0]$ and the starting direction at $\vdir_i = [0,0,1]$, we  analyze the solution space of a simplified non-planar problem where we vary the ending point $\x_f = [x,0,z]$ and the ending direction $\vdir_f = [\sin(\phi), \cos(\phi),0]$ for values of $x, z \in [-6,6]$ and $\phi \in [0,2\pi)$. The initial guess for the solver is set at $h_i = h_f = 0$.
Figure~\ref{fig:3dsweep} shows the results.

Again, the produced solution space shows expected symmetries.
When fixing $x = 0$, the solution space has vertical ``stripes'' to indicate that changing the angle $\phi$ does not change the number of solutions (Figure~\ref{fig:3dsweep}(a)). 
Additionally, fixing some value of $z,$ (in this case $z = -1$) makes  the number of solutions symmetric about $\phi = \pi$ since the relative alignment between the starting and ending directions would be equivalent (Figure~\ref{fig:3dsweep}(b)).
Finally,  the number of solutions found  is rotationally symmetric about the starting position. 
That is, if we fix $\phi = \phi_1$ for $\vdir_i = [\sin(\phi_1), \cos(\phi_1),0]$ and vary $\x_f = [x,0,z]$,  the landscape of the number of solutions for the slice where $\phi = \pi - \phi_1$ is a mirror image (Figure~\ref{fig:3dsweep}(c-d)). 

Similar to the planar case, the outer edges of the plots,  corresponding to the final configuration being farther away, produce in many cases the four regular solutions for $h_i$ and $h_f$.
We  also observe that non-planar cases with a  sufficiently small distance between initial and final configurations  result in 5 solutions, which must contain a switched solution. This behavior cannot be achieved by planar cases, since it is limited to four possible combinations between the circles.

\subsection{Varying Starting Seeds}
\begin{figure}[tb]
    \centering
    \includegraphics[width=0.496\textwidth]{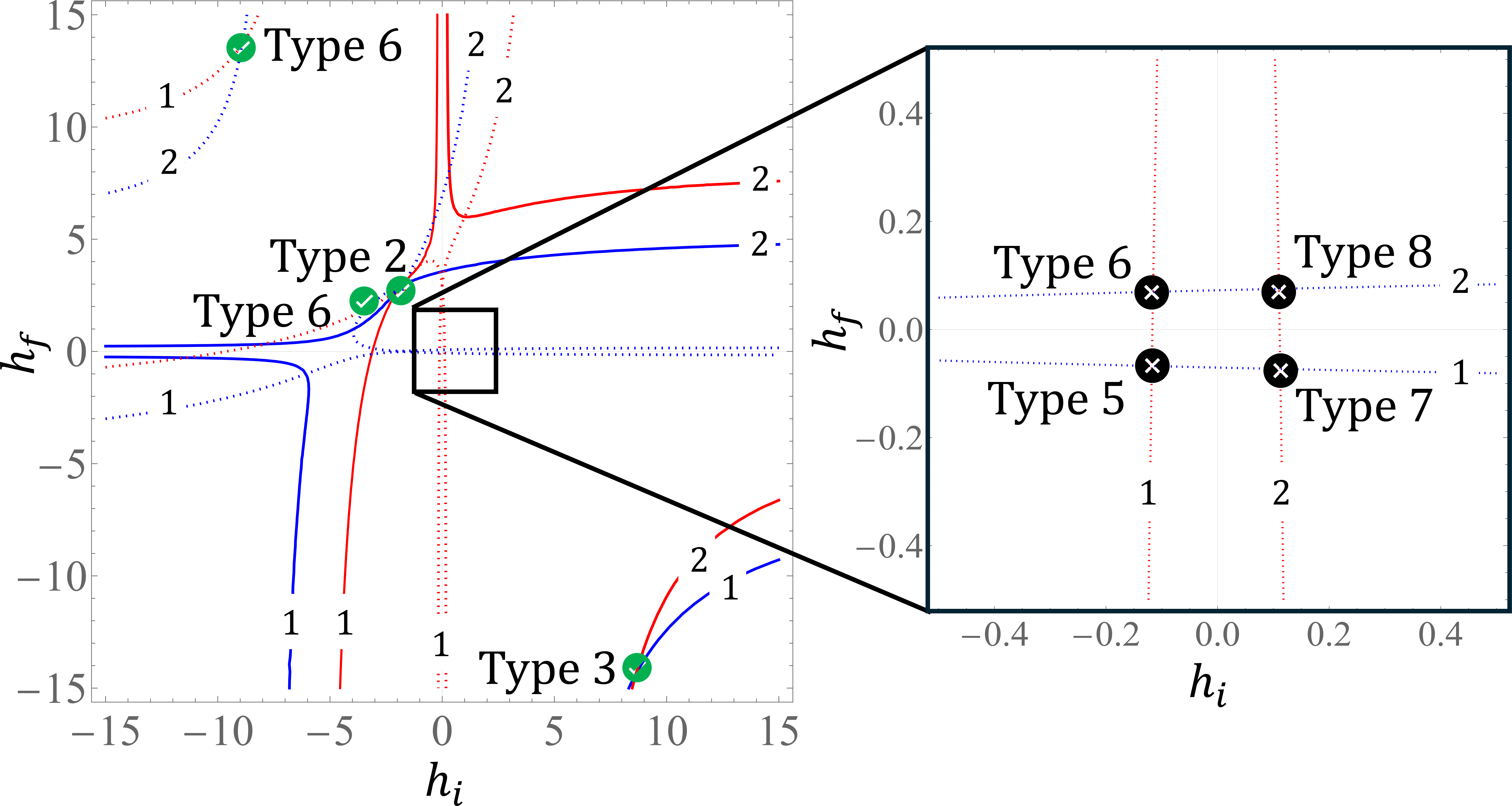}
    \includegraphics[width=0.496\textwidth]{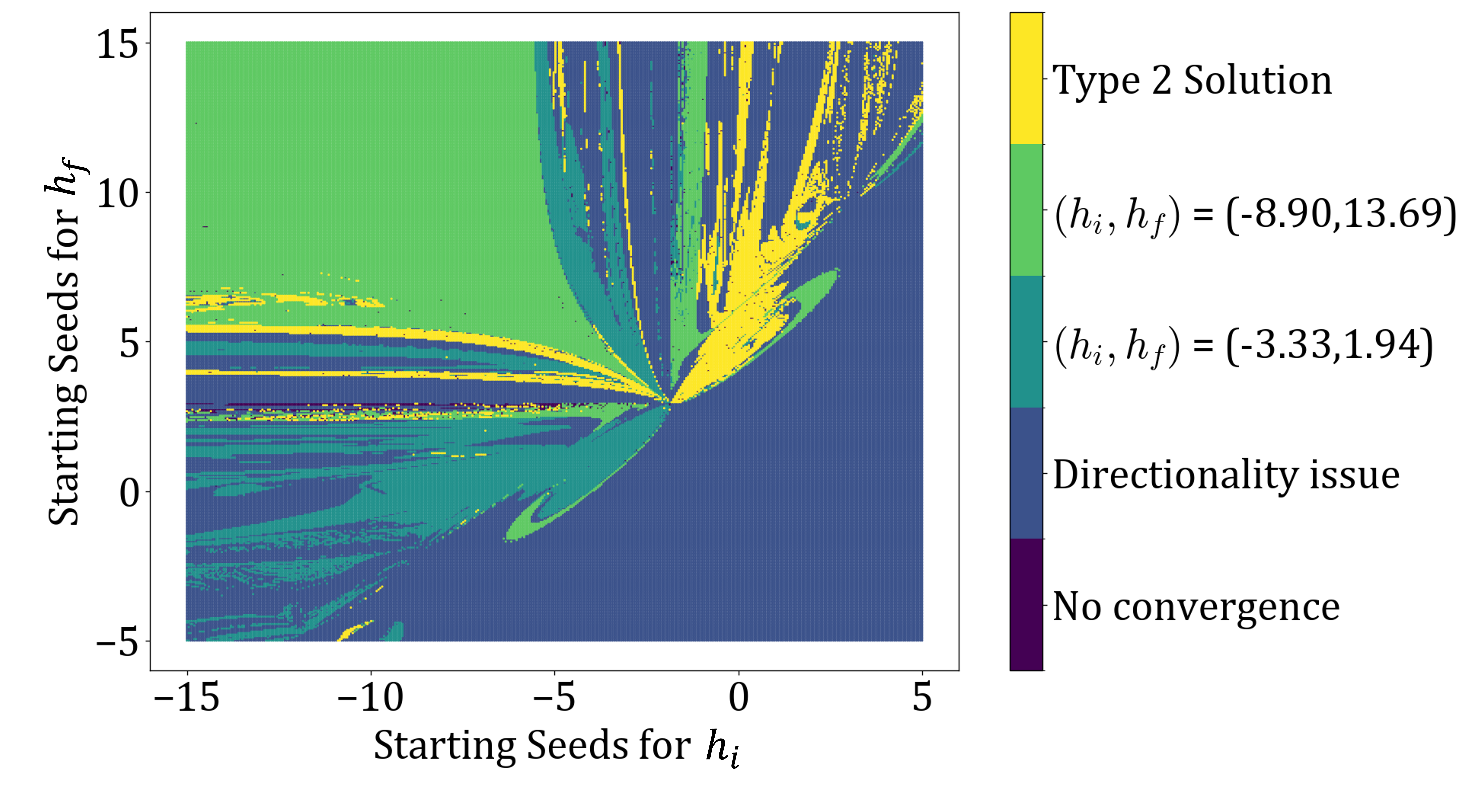}
    \caption{\textbf{Effect of Initial Starting Seeds} (Left) Solution plot for $\x_i = [0,0,0], \x_f = [1.059, 0, -4.588], \vdir_i = [0,0,1], \vdir_f = [-0.361,0,0.932]$ with multiple type 6 solutions. (Right) Solver convergence behavior for the Type 6 solution, depending on initial $h_i, h_f$ seeds. Yellow areas converge to the Type 2 solution $h_i = -1.804, h_f = 3.004$ that geometrically looks very close to a Type 6.}
    \label{fig:special_case}
\end{figure}

In some cases, there are multiple solutions for a single solution type. Using a numerical solver, at most one of these solutions will be found, and
the performance of the solver depends on the starting seed. 
We investigate this effect on a case
where the starting and ending points are close enough that the valid $h_i, h_f$ values are not the four regular solutions. When $\x_i = [0,0,0], \x_f = [1.059, 0, -4.588], \vdir_i = [0,0,1], \vdir_f = [-0.361,0,0.932]$, there are two switched type 6 solutions ($p_i^* > h_i, p_f^* < h_f$). 
Figure~\ref{fig:special_case} shows which solutions are found for different starting seed values.
We find that when varying the initial guesses for $h_i$ and $h_f$, the solver is able to converge to one of the intersection points in almost all cases.
However, the solution that the solver converges to is incredibly sensitive to the starting seed, indicating that it may be difficult to force the solver to find a different solution, even if it is known that others exist.

\subsection{Performance Comparison}
\subsubsection{Efficacy of Gradient Inclusion in Solving for Solutions}
\begin{figure}[tb]
    \centering
    \includegraphics[width=0.9\textwidth]{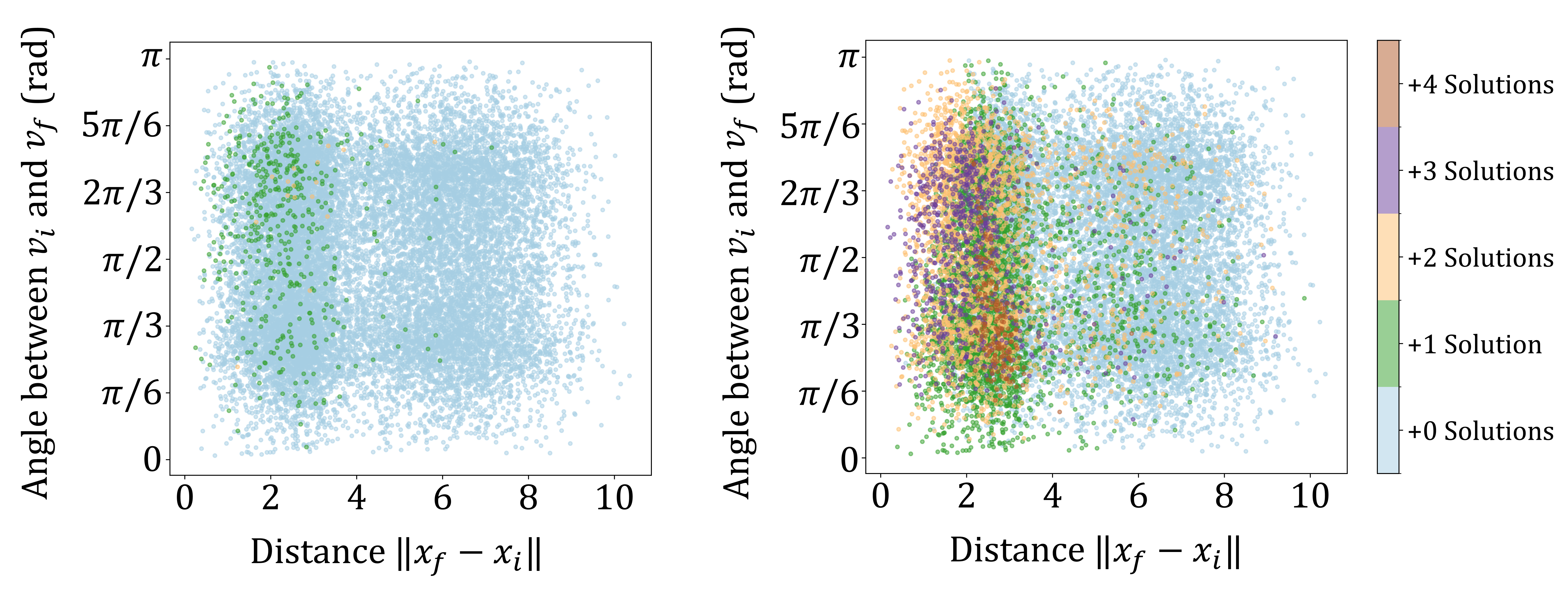}
    \caption{\textbf{Effect of Gradients on Numerical Solver and Comparison to~\cite{hota2010}} (Left) Number of additional solutions found when using \texttt{fsolve} with minus without the gradient for 18,000 random cases. (Right) Number of solutions found when using \texttt{fsolve} with our formulation minus \cite{hota2010} for 18,000 random cases.
    }
    \label{fig:gradient_efficacy}
\end{figure}

Our equations admit gradients, which may provide the solver with some assistance in converging to a valid solution.
We run a test with $\x_i = [0,0,0], \vdir_i = [0,0,1]$ and 18,000 random final configurations and initial guesses for $h_i$ and $h_f$.
For each final configuration and initial guess combination, we compute the CSC path solutions with and without the gradients and compare the number of solutions found (Figure~\ref{fig:gradient_efficacy}).
%
%
In the majority (95\%) of cases, the solver is able to find the same solutions with and without the gradient. However, when the starting and ending configurations are closer together with large orientation differences, the gradient makes the calculation easier, helping the solver to find up to 2 more solutions. 
This makes sense since the closer the initial and final positions are, the more influence the alignment of the directions will have over orientations of the arc and straight line components. 

\subsubsection{Comparison with Existing Method~\cite{hota2010}}
Finally, we analyze how solving these equations compares to existing solution methods. We refer to \cite{hota2010}, which is a different geometrical derivation for \threeD CSC Dubins paths resulting in a nonlinear system of equations over five variables.
We input the associated equations into \texttt{fsolve} and 
 run the solver on the same 18,000 test cases as above and compare the number of solutions found (Figure~\ref{fig:gradient_efficacy}).
In all cases, the number of solutions found using our simplified formulation is at least as many as that found using~\cite{hota2010}, and in the large majority of cases (95.7\%), our formulation is able to produce more solution paths: 4.39\% of the time we find the same number of solutions, 13.63\% of the time we find one more solution, 21.31\% of the time we find two more solutions, 59.65\% of the time we find three more solutions, and 1.07\% of the time we find four more solutions. 
We observe that our formulation produces a larger number of valid solutions when the distance between the initial and final configuration is small and, in some cases, the formulation in~\cite{hota2010} is unable to produce any solutions at all.
We therefore conclude that our simplified formulation is able to streamline and robustify the computation of CSC paths.

Note that for this test, we did not vary the initial seeds, and the quality of the output may change if multiple solution attempts are made. 
The time taken to solve each system of equations, for our formulation (with and without gradients) or for \cite{hota2010}, was approximately 0.01~s per test case.

\section{Conclusion}\label{sec:conclusion}

We investigate the path planning problem for \threeD Dubins vehicles, focusing particularly on  CSC paths, and we present a parameterization of the problem for computing solution paths by solving a system of 2 nonlinear equations. We have shown that these equations can be written explicitly from the input positions and orientations, and that we can  write analytical derivatives for these equations. We have used these equations together with the off-the-shelf Python \texttt{fsolve} solver and present results for multiple valid paths for a number of inputs. Further, we have demonstrated the ability to find more valid solutions than an existing geometric method~\cite{hota2010}, even at small distances between initial and final configurations, a limitation that other approaches  have encountered. 
When the starting and ending positions are far apart, almost all solution types produce valid paths. However, when the starting and ending positions are close, some solution types do not produce valid paths, and others produce multiple valid paths.
We have started to characterize empirically when these different situations occur. Future work includes a more rigorous study of the solution space for this problem.

We observe that our empirical results highly depend  on the behavior of the numerical solver being used. In particular, for solution types that have multiple valid solutions, the generated output depends on the initial seed for the solver. In this paper, we used the \texttt{fsolve} solver, which may produce outputs that are not the closest solution (in terms of $h_i,h_f$ distance) to the initial seed. 
However, at the same time, we were able to compute analytical derivatives for the equations that we used. We therefore expect that these gradients can be used in gradient-descent based algorithms to search for solutions that are close to an initial guess and reduce variability in the output. These approaches may be useful for future work in real-time planning for \threeD vehicles, where small adjustments to the current path may be preferred over large changes. Future work includes further investigation into improving consistency of output solutions for these scenarios.

\begin{credits}
\subsubsection{\ackname}
Support for this project has been provided in part by NSF Grant No. 2322898, by the Army Research Office under the SLICE Multidisciplinary University Research Initiatives Program award under Grant \#W911NF1810327, and by the AFOSR Multidisciplinary University Research Initiatives Program HyDDRA.
We also thank Wei-Hsi Chen and Daniel Feshbach for helpful discussions.

\subsubsection{\discintname}
The authors have no competing interests to declare that are
relevant to the content of this article.
\end{credits}
%
%
%
 \bibliographystyle{splncs04}
 \clearpage
 \bibliography{references}
\def\basetitle{Reparametrization of \threeD CSC Dubins Paths Enabling \twoD Search}
%
%
\author{Ling Xu\inst{1} \and
Yuliy Baryshnikov\inst{2} \and
Cynthia Sung\inst{1}}
\authorrunning{L. Xu et al.}
%
\institute{General Robotics, Automation, Sensing \& Perception Lab,\\
University of Pennsylvania, Philadelphia, PA, USA\\
\email{\{xu5,crsung\}@seas.upenn.edu}
\and
Department of Mathematics\\
University of Illinois Urbana-Champaign, Urbana, IL, USA\\
\email{ymb@illinois.edu}
}
\title{Supplementary Materials:\\\basetitle}
\titlerunning{Supplementary Materials: \threeD CSC Dubins Paths}
\maketitle 

\renewcommand{\thesection}{\Alph{section}}
\renewcommand{\thefigure}{\Alph{figure}}

\section{Example Solutions}\label{app:examples}

This section includes additional test cases to support the discussion in Section~\ref{sec:results_examples}

\subsection{Planar Cases}

\begin{figure}[tb]
\centering
      \includegraphics[width=\textwidth]{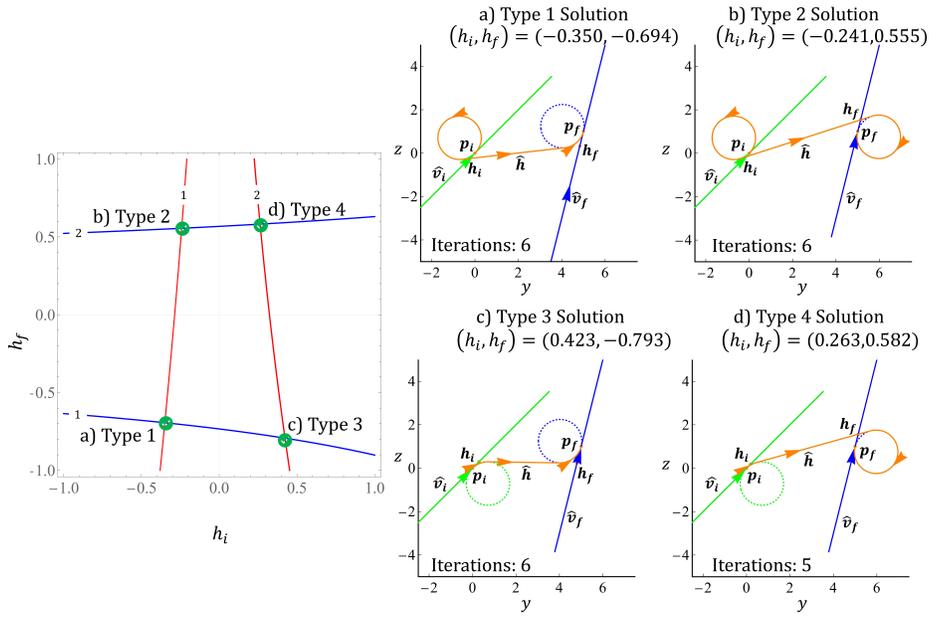}
      \caption{\textbf{\twoD  Problem Case 2 (Far)}:  $\x_i = [0,0,0], \x_f = [0, 5, 1],  \vdir_i = \frac{1}{\sqrt{2}}[0,1,1], \vdir_f = \frac{1}{\sqrt{17}}[0,1,4]$.  (Left) Solution plot for regular solutions of $p_i$ (red) and $p_f$ (blue). Intersections marked by green checks show the $(h_i, h_f)$ solution pairs. (Right) CSC paths for each solution type 1-4 marked on the left.}
   \label{fig:Ng2} \label{fig:planarfar2}
\end{figure}

Figure~\ref{fig:planarfar2} shows another planar test case where the starting and ending positions are far: $\x_i = [0,0,0], \x_f = [0, 5, 1], r = 1, \vdir_i = \frac{1}{\sqrt{2}}[0,1,1], \vdir_f = \frac{1}{\sqrt{17}}[0,1,4]$.
Four valid regular solutions were found in 0.034~s, with fewer than 10 solver iterations required for each solution type.

\label{app:planarclose2}
Figure~\ref{fig:planarclose2} shows another planar test case where the starting and ending positions are close: $\x_i = [0,0,0], \x_f = [1.8, 0, 3], \vdir_i = [0,0,1], \vdir_f = [0,0,-1]$.
This test case results in 2 regular solutions and 2 additional switched solutions. The solver found these paths in 0.101~s and required up to 18 iterations for each solution type.

\begin{figure}[tb]
\centering
      \includegraphics[width=\textwidth]{figs/figure7finalpic.png}
    \caption{\textbf{\twoD Problem Case 2 (Close)}:  $\x_i = [0,0,0], \x_f = [1.8, 0, 3], \vdir_i = [0,0,1], \vdir_f = [0,0,-1]$. (Left) 
    Solution plot for regular solutions of $p_i$ (solid red), regular solutions of $p_f$ (solid blue), switched solutions of $p^*_i$ (dotted red), and switched solutions of $p^*_f$ (dotted blue). 
    Intersections marked by green checks show the $(h_i, h_f)$ valid solution pairs,
    and
    black xs show invalid solutions. 
    (Right) CSC paths for each solution type marked on the left.}
    \label{fig:planarclose2}
\end{figure}

\subsection{Non-Planar Cases}
\begin{figure}[tb]
\centering
      \includegraphics[width=\textwidth]{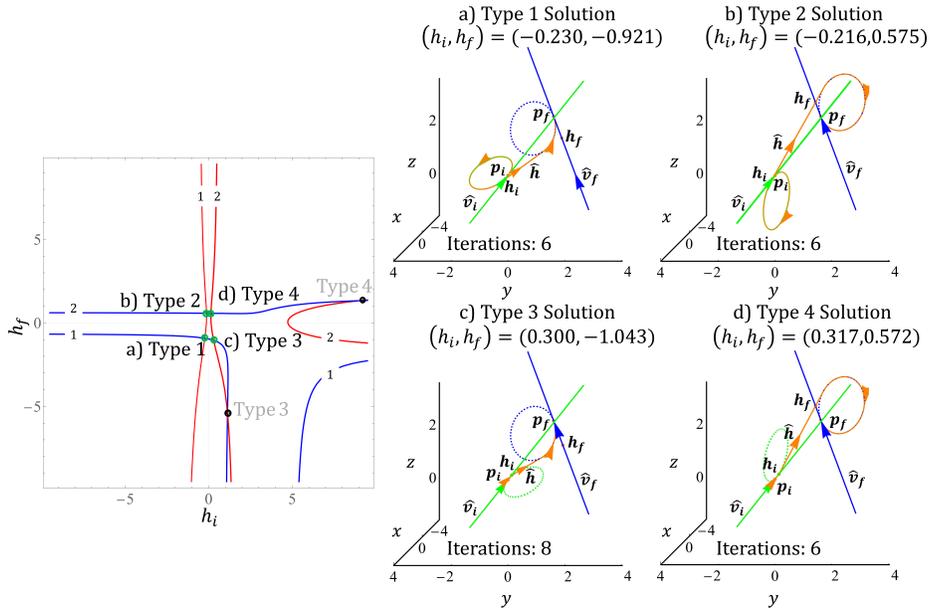}
    \caption{\textbf{\threeD Problem Case 2 (Far)}: $\x_i = [0,0,0], \x_f = [1, 2, 2], \vdir_i = \frac{1}{3}[1,1,1], \vdir_f = \frac{1}{43}[3,-3,5]$. 
     (Left) Solution plot for regular solutions of $p_i$ (red) and $p_f$ (blue). Intersections marked by green checks show the valid $(h_i, h_f)$ solution pairs.  There are extra Type 3 and Type 4 solutions that are invalid, marked by black xs. (Right) CSC paths for each solution type 1-4 marked on the left.
     }
\label{fig:3dfar2}
\end{figure}

Figure~\ref{fig:3dfar2} shows a second non-planar test case where the starting and ending positions are far: $\x_i = [0,0,0], \x_f = [1, 2, 2], r = 1, \vdir_i = \frac{1}{3}[1,1,1], \vdir_f = \frac{1}{43}[3,-3,5]$.
Four valid regular solutions were found in 0.040~s, with fewer than 10 solver iterations required for each solution type.

Figure~\ref{fig:3dclose2} shows a second example of a non-planar case where starting and ending points are close:
$\x_i = [0,0,0], \x_f = [1, -0.5, 2], \vdir_i = \frac{1}{\sqrt{41}}[-2,1,-6], \vdir_f = \frac{1}{2}[1,0,1]$. For this case, the solver finds solutions of Types 1, 2, 5, and 6. The solver took 0.109~s to find all the solutions, with a range of 11-30 iterations required for each solution type. 

\begin{figure}[tb]
\centering
      \includegraphics[width=\textwidth]{figs/figure11finalpic.png}
    \caption{\textbf{\threeD  Problem Case 2 (Close)}: $\x_i = [0,0,0], \x_f = [1, -0.5, 2], \vdir_i = \frac{1}{41}[-2,1,-6], \vdir_f = \frac{1}{2}[1,0,1]$.  
     (Left) 
    Solution plot for regular solutions of $p_i$ (solid red), regular solutions of $p_f$ (solid blue), switched solutions of $p^*_i$ (dotted red), and switched solutions of $p^*_f$ (dotted blue). 
    Intersections marked by green checks show the $(h_i, h_f)$ valid solution pairs,
    and
    black xs show invalid solutions. 
    There are 2 regular, 2 switched, and 3 invalid (Type 3, 5 7) solutions. 
     (Right) 
    CSC paths for each of the solution types marked on the left.
    }
  \label{fig:3dclose2}
\end{figure}
\end{document}